%% file: main.tex
\documentclass[10pt,twocolumn,letterpaper]{article}

\usepackage{template}
\usepackage{times}
\usepackage{epsfig}
\usepackage{amsthm}

\theoremstyle{definition}

\usepackage{graphicx}
\usepackage{amsmath}
\usepackage{amssymb}
\usepackage{booktabs}
\usepackage{balance}
\usepackage{multirow}
\usepackage{makecell}
\usepackage{microtype}
\usepackage{subfigure}
\usepackage{enumerate}
\usepackage[pagebackref=true,breaklinks=true,letterpaper=true,colorlinks,bookmarks=false]{hyperref}
\usepackage{authblk}
\usepackage[capitalize]{cleveref}
\crefname{section}{Sec.}{Secs.}
\Crefname{section}{Section}{Sections}

\usepackage{floatrow}
\newcommand{\sysname}{ElasticViT}

\iccvfinalcopy
\ificcvfinal\fi

\begin{document}

\newcommand{\lz}[1]{{\textcolor{red}{\it Lyna: #1}}}	
\newcommand{\tc}[1]{{\textcolor{blue}{\it Chen: #1}}}
\newcommand{\ting}[1]{{\textcolor{green}{\it Ting: #1}}}



\title{\sysname: Conflict-aware Supernet Training for Deploying Fast Vision Transformer on Diverse Mobile Devices}

\author{Chen Tang$^{1*\mathsection}$ \hspace{5pt} Li Lyna Zhang$^{2*\ddagger}$ \hspace{5pt} Huiqiang Jiang$^2$ \hspace{5pt} Jiahang Xu$^2$ \hspace{5pt} Ting Cao$^2$  \newline Quanlu Zhang$^2$\hspace{5pt}  Yuqing Yang$^2$\hspace{5pt}  Zhi Wang$^1$ \hspace{5pt} Mao Yang$^2$\\\fontsize{10}{10} \selectfont{$^1$Tsinghua University, $^2$Microsoft Research}}

\maketitle
\def\thefootnote{*}\footnotetext{Equal contribution}
\def\thefootnote{$\mathsection$}\footnotetext{Work is done during the internship at Microsoft Research}
\def\thefootnote{$\ddagger$}\footnotetext{Corresponding author, email: lzhani@microsoft.com}
\def\thefootnote{}\footnotetext{}
\ificcvfinal\thispagestyle{empty}\fi

 \input{abstract}
 \input{intro}

 \input{rw}

 \input{searchspace}
 \input{method1}
\input{eval}

 \input{conclusion}
\balance
{\small

\bibliographystyle{ieee_fullname}
\bibliography{ref}
}

\end{document}

%% file: abstract.tex
\begin{abstract}
Neural Architecture Search (NAS) has shown promising performance in the automatic design of vision transformers (ViT) exceeding 1G FLOPs. However, designing lightweight and low-latency ViT models for diverse mobile devices remains a big challenge. In this work, we propose {\sysname}, a two-stage NAS approach that trains a high-quality ViT supernet over a very large search space that supports a wide range of mobile devices, and then searches an optimal sub-network (subnet) for direct deployment. However, prior supernet training methods that rely on uniform sampling suffer from the gradient conflict issue: the sampled subnets can have vastly different model sizes (e.g., 50M vs. 2G FLOPs), leading to different optimization directions and inferior performance. To address this challenge, we propose two novel sampling techniques: complexity-aware sampling and performance-aware sampling. Complexity-aware sampling limits the FLOPs difference among the subnets sampled across adjacent training steps, while covering different-sized subnets in the search space. Performance-aware sampling further selects subnets that have good accuracy, which can reduce gradient conflicts and improve supernet quality. Our discovered models, {\sysname} models, achieve top-1 accuracy from 67.2\% to 80.0\% on ImageNet from 60M to 800M FLOPs without extra retraining, outperforming all prior CNNs and ViTs in terms of accuracy and latency. Our tiny and small models are also the first ViT models that surpass state-of-the-art CNNs with significantly lower latency on mobile devices. For instance, {\sysname}-S1 runs 2.62$\times$ faster than EfficientNet-B0 with 0.1\% higher accuracy.

\end{abstract}

%% file: intro.tex
\vspace{-3ex}
\section{Introduction}
\vspace{-1ex}
Vision Transformers (ViTs) have achieved remarkable success in various computer vision tasks~\cite{vit,swin,deit,detr,deformabledetr}.
However, the success comes at a significant cost - ViTs are heavy-weight and have high inference latency costs, posing a great challenge to bring ViTs to  resource-limited mobile devices~\cite{vitstudy}. Designing accurate and low-latency ViTs  becomes an important but challenging problem. 

Neural Architecture Search (NAS) provides a powerful tool for automating efficient DNN design. Recently, two-stage NAS such as BigNAS~\cite{bignas} and AutoFormer~\cite{autoformer}, decouple training and searching process and achieves remarkable search efficiency and accuracy. The first stage  trains a weight-shared supernet assembling all candidate architectures in the search space, and the second stage uses typical search algorithms to find best sub-networks (subnets) under various resource constraints. The searched subnets can directly inherit supernet weights for deployment, achieving comparable accuracy to those retrained from scratch. Such two-stage NAS can eliminate  the prohibitively expensive cost for traditional NAS to  retrain each subnet, making it a practical approach for efficient deployment.

\begin{figure}[t]
	\centering
	\includegraphics[width=1\columnwidth]{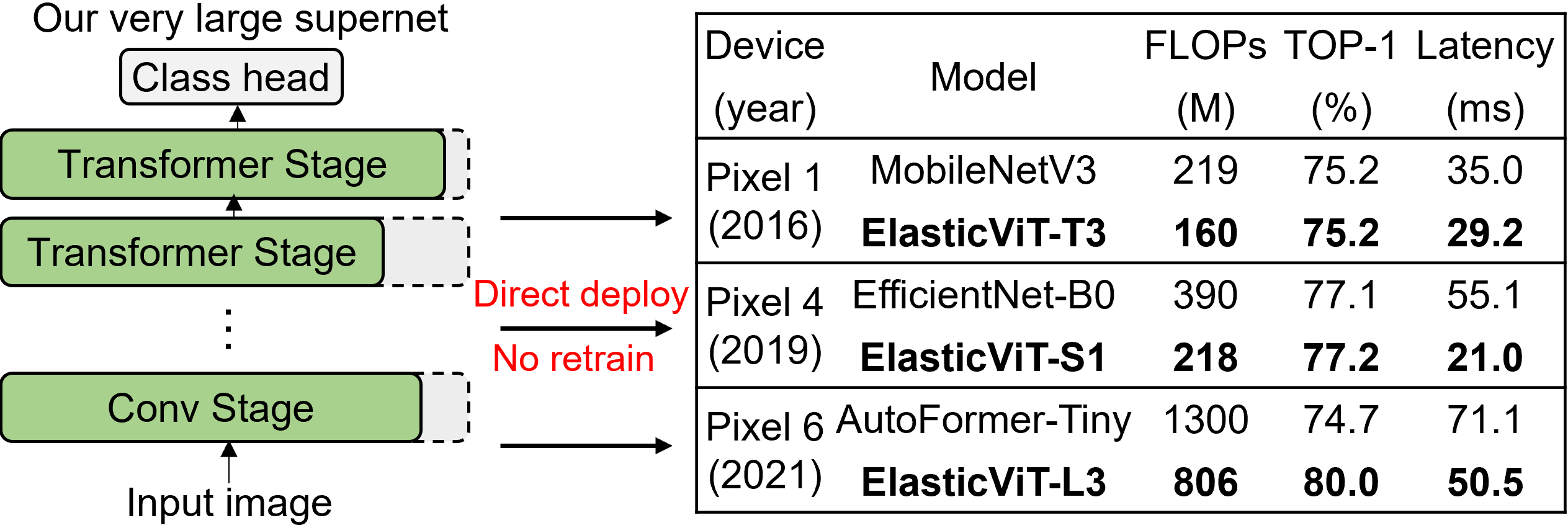}	
	\vspace{-4.5ex}
	\caption{We train a high-quality ViT supernet for a wide range of mobile devices. Our discovered ViTs outperform SOTA CNNs and ViTs with higher accuracy, fewer FLOPs and faster speed. }
	\label{fig:teaser}
\end{figure}

The success of two-stage NAS heavily relies on the quality of the supernet training in the first stage. However, it's extremely challenging to train a high-quality ViT supernet
for mobile devices, due to the  \textbf{vast mobile diversity}: mobile applications must support a wide range of mobile phones with varying computation capabilities, from the latest high-end devices to older ones with much slower CPUs. For instance, Google Pixel 1 runs $~$4$\times$ slower than Pixel 6. As a result, the supernet must cover ViTs that range from  tiny size ($<$ 100M FLOPs) for weak devices to large size for strong ones.
 However, including both tiny and large ViTs results in an overwhelmingly larger  search space compared to typical search spaces in two-stage NAS~\cite{ofa,bignas,nasvit}. Training a supernet over such a search space has been known to suffer from performance degradation due to optimization interference caused by subnets with vastly different sizes~\cite{nse,Yu2020Evaluating,zhang2020you}. 
While existing works~\cite{lin2020mcunet,hurricane,autoformer,su2022vitas} circumvent this issue by manually designing multiple separate normal-sized search spaces, the multi-space  approach can be costly. Moreover, there has been limited discussion regarding the root causes of this problem and how to effectively address it.

In this work, we introduce {\sysname}, a novel approach for training a high-quality vision transformer supernet that can efficiently serve both strong and weak mobile devices. Our approach is built upon  a single very large search space optimized for mobile devices, containing  a wide range of vision transformers with sizes ranging from 37M to 3G FLOPs. This search space is 10$^7\times$ larger than typical two-stage NAS search spaces, allowing us to accommodate a broad range of mobile devices with various resource constraints.


We start by investigating the root causes of poor performance when training a ViT supernet over our excessively large search space. 
We found that the main reason  is that prior supernet training methods rely on uniform sampling~\cite{autoformer,nasvit,bignas,ofa}, which can easily sample subnets with vastly different model sizes (e.g., 50M vs. 1G FLOPs) from our search space. This leads to conflicts between subnets' gradients and creates optimization challenges. We make two key observations:\textit{(i) the gradient conflict between two subnets increases with the FLOPs difference between them;} and \textit{(ii) gradient conflict between same-sized subnets can be significantly reduced if they are good subnets.}


Inspired by the above observations, we propose two key techniques to address the gradient conflict issue.  First, we propose  \textit{complexity-aware sampling} to limit the difference in FLOPs between sampled subnets across adjacent training steps, while ensuring that different-sized subnets within the search space are sampled. We achieve this by constraining the FLOPs level of the sampled subnets to be close to that of the previous step. Furthermore, we employ a multiple-min strategy to sample the nearest smallest subnet based on the FLOPs sampled at each step, thus ensuring performance bounds without introducing a large FLOPs difference with other subnets.
 Second, we introduce \textit{performance-aware sampling} that further reduces the gradient conflicts among subnets with similar FLOPs. Our method samples subnets with higher potential accuracy at each step from a prior distribution that is dynamically updated based on an exploration and exploitation policy. The policy leverages a memory bank and a ViT architecture preference rule. The preference rule guides the exploration of subnets with wider width and shallower depth, which are empirically preferred by ViT architectures. The memory bank stores historical  good subnets for each FLOPs level using prediction loss as a criterion.


Our contributions are summarized as follows:
\begin{itemize}
	\vspace{-1ex}
	\item We propose {\sysname} to automate the design of accurate and low-latency ViTs for diverse mobile devices. For the first time we are able to train a high-quality ViT supernet over a vast and mobile-regime search space.
		\vspace{-1ex}
	\item We conduct thorough analysis on the poor-quality supernet trained by existing approaches, and find that uniform sampling results in subnets of vastly different sizes, leading to gradient conflicts. 
	\vspace{-1ex}
	\item Inspired by our analysis, we propose two methods, complexity-aware sampling and performance-aware sampling, to effectively address the gradient issues by sampling good subnets and limiting their FLOPs differences across adjacent training steps.
	\vspace{-1ex}
	\item Extensive experiments on ImageNet~\cite{imagenet}
	and four mobile devices demonstrate that our discovered models achieve significant improvements over SOTA efficient CNN and ViT models in terms of both inference speed and accuracy. For example,  {\sysname}-T3 achieves
	 the same accuracy of 75.2\% as MobileNetV3 with only 160 MFLOPs, while be 1.2$\times$ faster. This is the first time that ViT outperforms CNN with a faster speed on mobile devices within the 200 MFLOPs range, to the best of our knowledge. {\sysname}-L achieves 80.0\% accuracy with 806 MFLOPs, which is 5.3\% higher than Autoformer-Tiny~\cite{autoformer} while using 1.61$\times$ fewer FLOPs. We also prove that {\sysname} substantially enhance the quality of supernet training, resulting in a noteworthy 3.9\% accuracy improvements for best-searched models. 
\end{itemize}

%% file: rw.tex
\vspace{-3ex}
\section{Related works}
\vspace{-1ex}
\noindent\textbf{Efficient Vision transformers}. 
 Many methods have been proposed to design efficient ViTs. They use different ways, such as new architectures or modules~\cite{chen2021crossvit,hassani2021escaping,li2022sepvit}, better attention operation~\cite{mobilevitv2,edgevit,swin,nextvit}  and hybrid CNN-transformer~\cite{wu2021cvt,cmt,mobilevit,mobilevitv3,levit,mobileformer,efficientformer}. Hybrid models usually perform well with small sizes by introducing special operations. For instance,  MobileFormer~\cite{mobileformer} uses a parallel CNN-transformer structure with bidirectional bridge. 
 However, although the FLOPs are reduced, these ViTs still have high latency because of  mobile-unfriendly operations such as the  bidirectional bridge.
  



\noindent\textbf{Neural Architecture Search}. NAS has
achieved an amazing success in automating the design of efficient CNN architectures~\cite{pham2018efficient,guo2020single,proxylessnas,bignas,ofa}. Recently, several works apply NAS to find improved ViTs, such as Autoformer~\cite{autoformer}, S3~\cite{autoformerv2}, ViTAS~\cite{su2021vitas} and ViT-ResNAS~\cite{liao2021searching}. These methods focus on searching models exceeding 1G FLOPs. For small ViTs, HR-NAS~\cite{ding2021hr}, UniNet~\cite{uninet} and NASViT~\cite{nasvit} search for hybrid CNN-ViTs and achieve promising results under small FLOPs. However, these NAS works mainly optimize for FLOPs without considering the efficiency on diverse mobile devices, which leads to suboptimal performance.




\noindent\textbf{Supernet Training}. Early NAS methods~\cite{zoph2016neural,zoph2018learning,real2019regularized,real2017large} are very costly because they need to train and evaluate many architectures from scratch. More recent one-shot NAS methods~\cite{guo2020single,proxylessnas,liu2018darts,chu2021fairnas} use weight-sharing to save time. But they still need to retrain the best architecture from scratch for higher accuracy, which is expensive when targeting multiple constraints.
To solve this problem, two-stage NAS, such as OFA~\cite{ofa}, BigNAS~\cite{bignas} and AutoFormer~\cite{autoformer} decouples the training and search. They train a supernet where the good subnets can be directly deployed without retraining. However, they employ uniform sampling to sample subnets,  which can lead to subnets with significantly different sizes being sampled in a much larger search space, resulting in gradient conflicts and inferior performance.
Our work proposes conflict-aware supernet training to address this issue.

%% file: searchspace.tex
\section{Search Space Design and Training Analysis}

	\vspace{-1ex}
\subsection{Search Space Design}
	\vspace{-1ex}


\noindent\textbf{Mobile-friendly ViTs}. While many works aim to design ViT models with high accuracy and small FLOPs, we observe that models with small FLOPs may have high latency on mobile devices. For example, NASViT-A0~\cite{nasvit} has fewer FLOPs than MobileNetV3, but it runs 2$\times$ slower on a Pixel4 phone.  MobileFormer~\cite{mobileformer} has only 52M FLOPs but runs 5.5$\times$ slower. This is because these models incorporate effective but \textit{mobile-unfriendly operations} that reduce FLOPs.

Our goal is to design accurate ViTs that can achieve low-latency on mobile devices. To achieve this, we draw inspiration from recent works~\cite{levit,nasvit} and construct our search space based on mobile-friendly CNN-ViT architecture as shown in  Table~\ref{tbl:searchspace}.
 In CNN stage, we use MobileNetv2~\cite{mobilenetv2} (MBv2) and MobileNetv3~\cite{mobilenetv3} (MBv3)  blocks. In ViT stage, we make key modifications  based on NASViT  attentions for better efficiency. We remove the slow talking heads~\cite{shazeer2020talking} and use Hswish instead of Gelu. We opt not to employ shifted window attention in Swin~\cite{swin}, because it does not help much when the input size is small. We measure the latency on real devices. Our attention can speed up latency by  $>$2$\times$, making the transformer block more efficient.

\vspace{2pt}
\noindent\textbf{A very large search space}.  
Unlike previous ViT NAS works~\cite{autoformer,su2021vitas} that focused primarily on large models exceeding 1G FLOPs, our search space must accommodate a wide range of ViT configurations, from tiny to large, to meet the demands of diverse mobile devices. For instance, a high-end device like the Pixel 6 can handle a large ViT model with 500M FLOPs to meet latency constraint of $\sim$30ms, whereas a less powerful device such as the Pixel 1 requires a tiny model with $<$100M FLOPs to meet the same constraint.


 Table ~\ref{tbl:searchspace} presents the final search space. We add many small choices for each block dimensions to include tiny ViTs. We also make several key designs to cover  potentially good subnets based on previous works. Specifically, we follow~\cite{park2022how} and increase the maximun width choices and decrease the depth choices for ViT stages. We follow LeViT~\cite{levit} and allow $V$ matrix to have larger expansion ratio of \{2, 3, 4\} (i.e., $V$ scale). We also allow larger expansion ratios of \{2, 3, 4, 5\} for MLP layers, because they are not redundant when using a typical ratio of 2.


In total, our search space covers a wide range of ViT subnets with varying sizes, from 37 to 3191 MFLOPs. It contains an enormous 1.09$\times10^{17}$ subnets, which is a significant increase of \textbf{10$^7$$\times$ larger} than a typical search space in two-stage NAS (refer to Appendix. C for details). This presents new challenges in training a high-quality supernet over such a large search space.



\begin{table}[t]
			\centering
		\fontsize{8.5}{8.5} \selectfont
			\vspace{-2.5ex}
		\caption{Our very large search space to support both weak and strong mobile devices. It's 10$^{7}\times$ larger than that in  two-stage NAS. Tuples of three values represent the lowest value, highest, and steps.}
		\label{tbl:searchspace}
		\begin{tabular}		{@{\hskip2pt}c@{\hskip2pt}c@{\hskip3pt}c@{\hskip3pt}c@{\hskip5pt}c@{\hskip5pt}c@{\hskip2pt}}
	\hline
			Stage&Depths&\makecell{Channels}&\makecell{Kernel size \\(V scale)}& \makecell{Expansion\\ ratio} & Stride\\
			\hline
			Conv 3$\times$3 & 1 & (16, 24, 8)& 3 & -&2\\
			MBv2 block& 1-2& (16, 24, 8) & 3, 5& 1 &  1\\
			MBv2 block& 2-5 & (16, 32, 8) & 3, 5& 3, 4, 5, 6&2\\
			MBv3 block & 2-6&(16, 48, 8) & 3, 5& 3, 4, 5, 6&2\\
			Transformer & 1-5&(48, 96, 16)&2, 3, 4&2, 3, 4, 5&2\\
			Transformer & 1-6&(80, 160, 16)&2, 3, 4&2, 3, 4, 5&1\\
			Transformer & 1-6&(144, 288, 16)&2, 3, 4&2, 3, 4, 5&2\\
			Transformer & 1-6&(160, 320, 16)&2, 3, 4&2, 3, 4, 5&2\\
			MBPool& -&1984&-&6&-\\
			\hline
			Input resolution & \multicolumn{5}{c}{128, 160, 176, 192, 224, 256}\\
		\hline
		\end{tabular}  
\end{table}

\subsection{Analysis of Training a Very Large Supernet}
\label{sec:analysis}
\vspace{-1ex}
\begin{figure*}[t]
	\centering
	\includegraphics[width=0.9\textwidth]{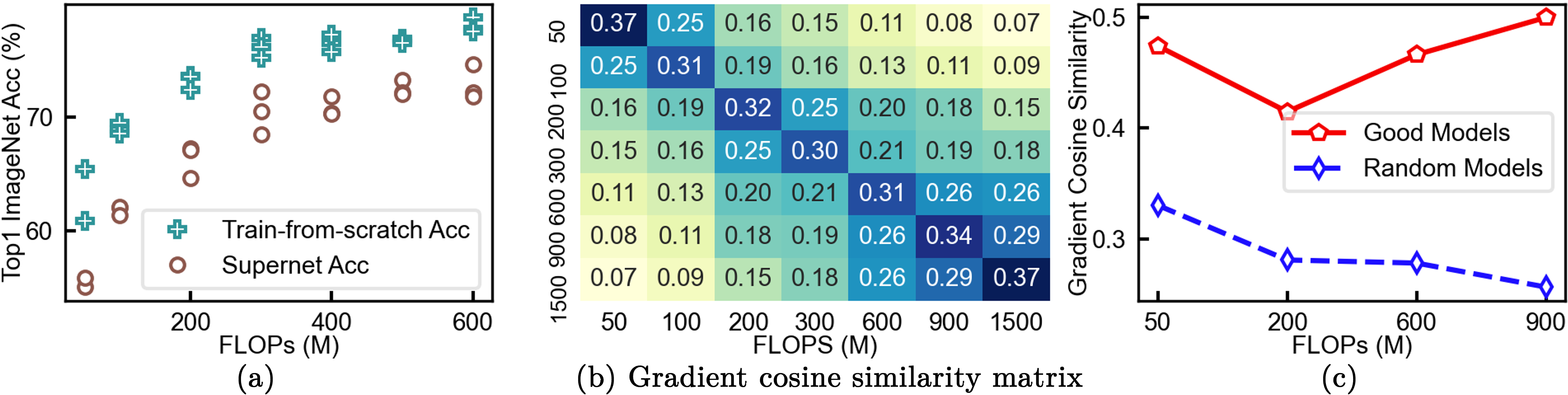}	
\vspace{-2ex}
	\caption{(a) Model accuracy achieved by training from scratch vs. inheriting supernet weights; (b) Gradient cosine similarity of models with different FLOPs; (c) Under the same FLOPs level, good models share more similar gradients than random sampled models.}
	\label{fig:gradient}
\end{figure*}
Supernet training differs from standard single network training in that all the subnets share weights for their common parts. The shared weights may receive conflicting gradients that lead to different optimization directions, which lowers the final accuracy. Several techniques have been proposed to mitigate this issue~\cite{bignas,alphanet,nasvit}, have demonstrated success in training high-quality supernet over typical search spaces. Among them, sandwich  rule is essential to ensure performance lower and upper bounds. Specifically, it samples  a min, max and two random subnets per iteration.

However, when training a ViT supernet over our vast  search space in Table~\ref{tbl:searchspace}, we observe significant accuracy drop using previous best practices~\cite{alphanet,nasvit}. Specifically, we use sandwich rule and follow the same training receipts in NASViT~\cite{nasvit}.  Fig.~\ref{fig:gradient}(a)  compares the accuracy of 20 random subnets achieved by inheriting supernet weights to retraining on ImageNet. Compared to retraining, models derived from the supernet experience an accuracy drop with up to 8\%.

\vspace{3pt}
\noindent\textbf{Analysis.} 
Compared to a typical ViT supernet, our supernet includes many tiny subnets and the sizes between subnets can differ greatly. 
We randomly sample subnets with MFLOPs ranging from 50 to 1200 from our supernet, and compute the cosine similarity of shared weights' gradient between each pair under the same batch of training data. A lower similarity indicates larger differences in gradients and thus more difficulty in training the supernet well. Fig.~\ref{fig:gradient}(b) shows that  \textit{gradient similarity of shared weights between two subnets is negatively correlated with their FLOPs difference} (\textbf{Observation\#1}). Subnets with similar FLOPs achieve the highest gradient similarity, while the similarity is close to 0 if there is a large FLOPs difference between two subnets.

Besides FLOPs difference, subnet quality may also affect gradient similarity of shared weights. If a poor subnet is sampled and trained, it would disturb the weights of good subnets. To verify this hypothesis, we randomly sample 50 subnets with same level FLOPs and compute the shared weights gradient cosine similarity between top subnets and randomly-sampled subnets. Fig.~\ref{fig:gradient}(c) suggests that \textit{gradient similarity between same-size subnets can be greatly improved if they are good subnets} (\textbf{Observation\#2}).

%% file: method1.tex
\section{Methodology}
Inspired by the observations in \cref{sec:preference_rule},  we propose two methods to address the gradient conflict issue. (i) We introduce  complexity-aware sampling
to restrict the FLOPs difference between adjacent training steps, as large difference incurs gradient conflict problem (\cref{sec:adjacent_sampling}). (ii) Since ``good subnets'' can enhance each other, which further reduce gradient conflicts, we propose performance-aware sampling to ensure the training process can sample good subnets as much as possible (\cref{sec:preference_rule}). 

\subsection{Preliminary}
For a search space $\mathcal{A}$, two-stage NAS aims to train a weight-sharing network (supernet) over $\mathcal{A}$ and jointly optimize the parameters of all subnets. 
Since it is infeasible to train all subnets in $\mathcal{A}$, this is often achieved by the widely-used \textit{sandwich rule}, which samples two types of subnets at each training step: \textbf{(i)} an expectation term approximated by a sampled subnet set through a prior distribution $\Gamma$ over the search space \{$s_m \vert s_m \sim \Gamma (\mathcal{A})$\}$_{m=1}^M$ and
\textbf{(ii)} a fixed subnet set $s_n \in \mathcal{S}$.  Formally, the supernet training  can be framed as the following optimization problem:
\begin{equation} 
	\small
	\begin{aligned}
		& \underset {w} { \operatorname {arg\,min} } \, \left[
		\mathbb{E}_{s_m \sim \Gamma(\mathcal{A})} \mathcal{L}_{D} \left( f \left( w_{s_m}\right) \right) + \sum_{s_n \in \mathcal{S}} \mathcal{L}_{D} (f(w_{s_n})) \right] \\
	\end{aligned} 
	\label{eq:sandwich_rule}
\end{equation} 
where $w$ is the shared weights for all subnets, $f(\cdot)$ denotes the neural network, $\mathcal{L}_\mathcal{T}$ is the loss on the training set $D$ and $w_s$ is the exclusive weights of subnet $s$.

Without loss of generality, there are two types of subnets in any space --  \emph{the random subnets that each dimensions are sampled between maximum and minimum  settings}; and \emph{the smallest and largest subnets that each dimensions	are the minimum and maximum settings, respectively}.
In recent works \cite{bignas, attentivenas, nasvit}, sandwich rule often approximates the first term in Eq. \ref{eq:sandwich_rule} by randomly sampling $M$=2 subnets from a uniform distribution $\Gamma$. In the second term, $\mathcal{S}$ typically consists of the largest subnet $s_l$ and the smallest subnet $s_s$.

We now revisit the effectiveness of applying sandwich rule on training our mobile-specialized supernet in Table~\ref{tbl:searchspace}. Obviously, it can easily cause gradient conflicts due to two reasons. First, it always sample the smallest (37M FLOPs), biggest (3191 MFLOPs) and 2 random subnets, which results in a significant FLOPs difference and often causes the gradient similarity to be close to 0 (Fig.~\ref{fig:gradient}(b)). Second, the size and quality of the 2 randomly sampled subnets cannot be guaranteed, which 
exacerbates the issue.


\vspace{-1ex}
\subsection{Complexity-aware Sampling}
\label{sec:adjacent_sampling} 
\vspace{-1ex}

In this section, we introduce complexity-aware\footnote{We use \rm{FLOPs} to represent the complexity metric, which can trivially be generalized to other metrics (e.g., latency).} sampling to mitigate the gradient conflicts by large FLOPs difference, while ensuring that different-sized subnets can be trained.

\noindent\textbf{Adjacent Step Sampling For Uniform Subnets.} 
Due to the extremely large space, randomly sampling $M$ subnets can result in significant FLOPs differences (\emph{see Appendix.C for detailed illustrations}) both within the same training step and across different training steps. 
Therefore, we propose to constrain the FLOPs of $M$ random subnets for each training step. 
As shown in Fig.~\ref{fig:alg}, we simply constrain the subnets within the same training step to have the same level of FLOPs.

However, it is non-trivial to effectively constrain the FLOPs differences between different training steps, as it requires ensuring that subnets of varying sizes can be trained to support diverse resource constraints. We propose  adjacent step sampling to gradually change the FLOPs level. 



Specifically, we define a set of gradually increased complexity levels ${C_1, ..., C_K}$ (e.g., {100, 200, ..., 800} {\rm MFLOPs}), which cover a range of ViT models from tiny to large. Suppose step $t$-1 samples the $i^{th}$ complexity level $C^{(t-1)}_i$, and step $t$ samples $M$ subnets $s^{(t)}$ under the $j^{th}$ complexity level of $C^{(t)}_j$. To satisfy the adjacent step sampling, the FLOPs distance between these two steps must satisfy the following:
\begin{equation}
	\small
	g\left(s^{(t)};  C^{(t-1)}_i\right) =  |C_j^{(t)} - C^{(t-1)}_i|=0
\end{equation}
To satisfy the equation, we offer $C^{(t)}_j$ three options: ${C_{i-1}^{(t-1)}, C_i^{(t-1)}, C_{i+1}^{(t-1)}}$, representing the choices of \emph{decreasing the FLOPs level by one}, \emph{maintaining the current FLOPs level}, or \emph{increasing the FLOPs level by one}, respectively.

\vspace{2pt}
\noindent\textbf{Remove the Biggest Subnet.} Since the biggest subnet has 3191 M\rm{FLOPs}  in our search space, it naturally introduces a large FLOPs difference with our considered complexity levels, which can cause gradient conflict at each step. 
Empirically, we find that removing the largest subnet stabilizes the training process and improves overall performance. 

\vspace{2pt}
\noindent\textbf{Use Multiple Hierarchical Smallest Subnets (HSS).} 
Since the smallest subnet has only 37 MFLOPs, sampling at the large complexity range ($C_i \geq $ 500 MFLOPs) can also introduce large FLOPs difference at each step. Unlike the biggest subnet, the smallest subnet decides the performance lower bound of the whole space \cite{tang2022arbitrary}, which cannot be removed simply. Therefore, instead of sampling the min subnet with only 37 MFLOPs at each step, we sample a nearest min subnet from the \emph{hierarchical smallest subnets (HSS) set} $\hat{\mathbf{S}} = \{s_n\}_{n=1}^N$. 
HSS set includes $N$=3 pre-defined subnets with discrepant complexity. 
At step $t$, when sampling around a complexity level $C^{(t)}$, we select a subnet $s_n \in \hat{\mathbf{S}}$ whose complexity is closest to it as the smallest subnet, as in Eq. \ref{eq:hss_equation}. 

In our experiments, we conduct empirical analysis and select the $N$=3 smallest subnets as the HSS set $\hat{\mathbf{S}}$. As shown in Fig.~\ref{fig:alg}, these subnets include the original 37 MFLOPs subnet (min$_1$), as well as a 160 MFLOPs subnet (min$_2$) and a 280 MFLOPs subnet (min$_3$), which are sampled and added as the second and third smallest subnets, respectively.

\emph{Discussion for HSS set.} The multiple ``smallest'' subnets in HSS set logically partition the whole search space into $N$ hierarchical sub-spaces (Fig.~\ref{fig:alg}). This is differs from previous methods\cite{autoformer,autoformerv2}, which manually divide the space into separate sub-spaces and train them individually. The HSS set offers two advantages: \textit{(i)}: it enables unified weight-sharing across the entire space, allowing small subnets to benefit from large subnets. Moreover, it has been proven that incorporating small models into larger ones can significantly enhance small subnets' performance\cite{cai2021network}. \textit{(ii)}: It does not rely on any heuristic space re-design or strong relationship assumption between dimensions (\emph{e.g.,} linear correlation in \cite{autoformerv2}), which enhances the universality of our method.

\vspace{2pt}
\noindent\textbf{Optimization Objective.} With applying complexity-aware sampling, we reformulate Eq. \ref{eq:sandwich_rule}  as the follows:
\begin{small}
	\begin{equation}
		\begin{aligned}
			\underset {w} { \operatorname {arg\,min} } \, \left[\sum^M_{s_m^{(t)} \in \mathbf{U}} \mathcal{L}_{D} \left(f(w_{s_m^{(t)}})\right) + \sum_{s_n \in \hat{\mathbf{S}}} \sigma (s_n, C^{(t)}_j) \mathcal{L}_{D} \left( f(w_{s_n}) \right) \right], 
		\end{aligned}
	\end{equation} 
\end{small}
where $t$ denotes the current training step, 
 $\mathbf{U}$ is the stochastic subnet set containing $M$=3 uniform subnets, in which  each subnet has the FLOPs level of $C^{(t)}_j$ for step $t$: 
\begin{small}
\begin{equation}
	\mathbf{U} = \left\{s_m^{(t)} \vert s_m^{(t)} \sim \Gamma (\mathcal{A}) \: \& \: g'\left(s^{(t)}_m; C^{(t)}_j \right) == 0 \right\}_{m=1}^M, 
	\label{eq:subnetwork_set_sampling}
\end{equation}
\end{small}
and $\sigma(\cdot)$ selects the nearest smallest subnet from HSS: 
\begin{small}
\begin{equation}
	\sigma(s_n, C^{(t)}_j)= 
	\begin{cases}
		1&\mbox{if $s_n$ is the nearest min that is larger than $C^{(t)}_j$}\\
		0&\mbox{otherwise, }
	\end{cases}
	\label{eq:hss_equation}
\end{equation}
\end{small}
Note that we still use the notion $\Gamma$ since the prior distribution will be determined in \cref{sec:preference_rule}.

\subsection{Performance-aware Sampling}
\label{sec:performance_aware_sampling}
In ~\cref{sec:analysis}, we observe that top-performing subnets can further alleviate the gradient conflict issue. Inspired by this, we introduce a performance-aware sampling  to sample subnets with potentially higher accuracy.

Specifically, for the $M=3$ same-sized subnets in $\mathbf{U}$ (as defined in Eq.~\ref{eq:subnetwork_set_sampling}), we aim to sample from a new distribution that favors good subnets, rather than a uniform distribution with random performance. To achieve this, we propose an exploration and exploitation policy that constructs the new distribution based on quality-aware memory bank and path preference rule. The quality-aware memory bank is used to exploit historical good subnets with a probability of $q$, while preference rule  explores subnets with wider width and shallower depth with a probability of $1-q$.

\begin{figure}[t]
	\centering
	\includegraphics[width=1\columnwidth]{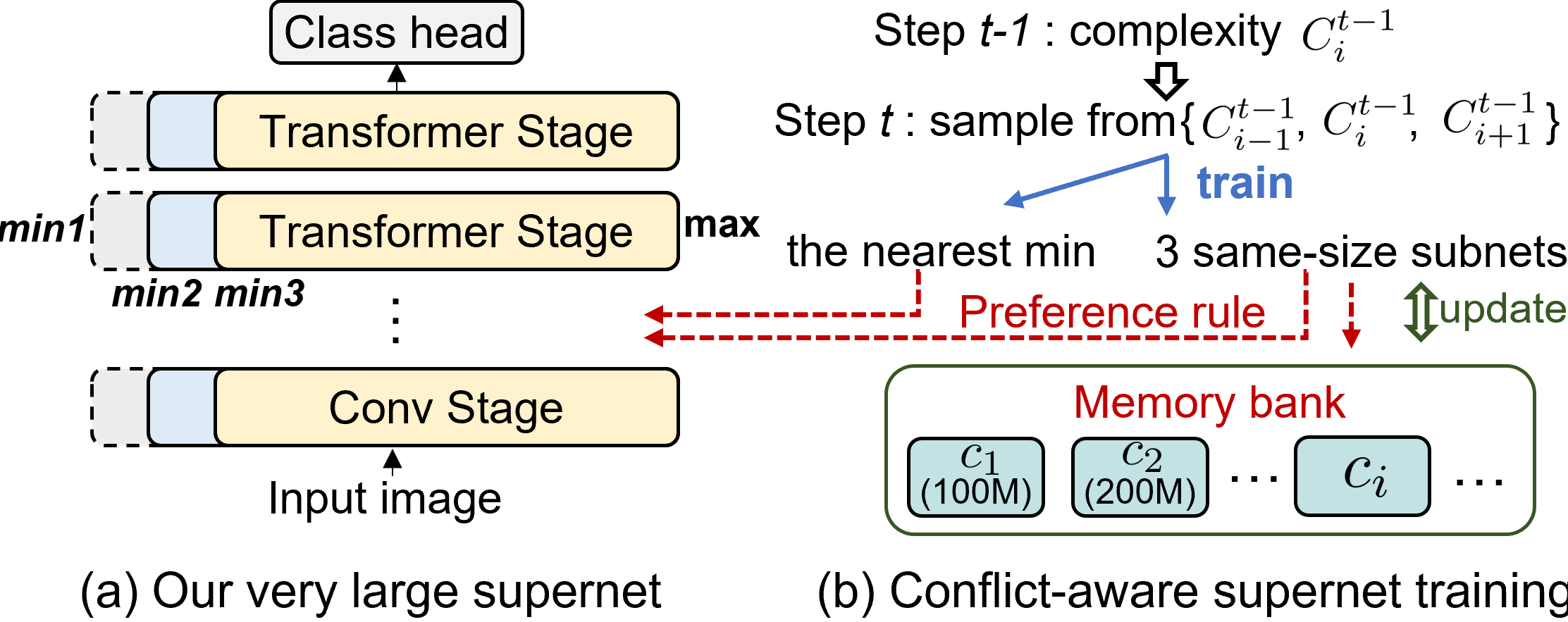}	
	\vspace{-4.5ex}
	\caption{The overview of our proposed confict-aware supernet training. At step $t$, we first sample a target FLOPs that is close to that at step $t$-1, then we sample 3 subnets with same level of FLOPs and find a nearest min subnet to update supernet weights.}
	\label{fig:alg}
\end{figure}

\vspace{2pt}
\noindent\textbf{Quality-aware Memory Bank.}
\label{sec:memory_bank}
As shown in Fig.~\ref{fig:alg}, \emph{memory bank} stores the historical up-to-date good subnets for each FLOPs level.  Specifically, the good subnets are identified through comparing cross-entropy loss on the current min-batch. Suppose  step $t$ sample the $j^{th}$ FLOPs level of $C_j$, 
 ${\rm B}_j$ denotes the good subnets  for FLOPs of $C_j$ at step $t$, 
the prior distribution $\Gamma$ can be as a mixture distribution of the dynamic memory bank and other unexplored subnets.
From this perspective,  the expectation of subnet $s^{(t)}_m$ is
\begin{gather}
	\mathbb{E}_{s_m^{(t)} \sim \Gamma (\mathcal{A})} = q \cdot U({\rm B}_j) + (1-q) \cdot U(\hat{\mathcal{A}}_{C_j}), \label{eq:quality_aware_sampling} \\ 
	\text{where \;} {\rm B}_j \cup \hat{\mathcal{A}}_{C_j} = \mathcal{A}_{C_j} \nonumber
\end{gather}
where $\mathcal{A}_{C_j}$ denotes all the subnets with FLOPs level of $C_j$ in search space $\mathcal{A}$.   $U(\hat{\mathcal{A}}_{C_j})$ is the uniform distribution of unexplored subnets, which will be further regulated by the path preference rule in next section.

At the early training, a relatively small value (i.e., 0.2) of $q$ is applied so that uniform sampling dominates the training to exploring promising subnets. 
As training proceeds, the memory banks are gradually filled, at which $q$ is also gradually increased to exploit these memorized good subnets. 

\emph{Memory bank replacement strategy.} 
We adopt the \emph{Worst-Performing} strategy to replace the subnet in the memory bank. When the current subnet outperforms the worst-performing subnet in the memory bank, the worst-performing subnet is replaced by the current subnet.

\vspace{2pt}
\noindent\textbf{Path Preference Rule.}
\label{sec:preference_rule}
In Eq.~\ref{eq:quality_aware_sampling}, when exploring unvisited subnets from $\hat{\mathcal{A}_{C_j}}$, we propose to sample ViT-preferred architectures that are more likely to achieve higher accuracy. Our approach is inspired by recent studies~\cite{park2022how, raghu2021vision}, which found that in pure ViT models, the later Transformer stages tend to prefer wider channels over more layers compared to CNN models. We empirically verify that this conclusion holds true in the hybrid CNN-Transformer NAS space (see Appendix), which motivates us to incorporate this preference into the super-network training to filter out inferior subnets.

Specifically, when a subnet $s^{(t)} = \{o_{width}^{(t)}, {o_{depth}^{(t)}\}}$ is sampled  (\emph{i.e.,} through uniform distribution of second term in Eq. \ref{eq:quality_aware_sampling}), we expect it to have \emph{wider widths and shallower depths in Transformer stages}, where $o_{width}^{(t)}$ and $o_{depth}^{(t)}$ are the width and depth dimension (\emph{i.e.,} the dimension of interests \footnote{For simplicity, we omit the remaining dimensions (e.g., kernel size, expansion ratio, etc.) since they are not the dimensions of interest.}), respectively. 

However, it's non-trivial to determine whether a ViT is a deep and narrow model or a shallow and wide model. Our  approach involves quantifying a subnet's FLOPs distribution in terms of the depth and width dimensions by comparing it with an anchor model. This enables us to determine the subnet's preference in terms of depths and widths. Specifically, suppose the anchor model is $s^{ac} = \{o_{width}^{ac}, {o_{depth}^{ac}\}}$ and the sampled subnet is 
$s^{(t)} = \{o_{width}^{(t)}, {o_{depth}^{(t)}\}}$, our first step is to generate a new subnet $\overline{s^{(t)}} = \{{\bf o_{width}^{ac}}, o_{depth}^{(t)}\}$ by aligning all dimensions to anchor model except the depth dimension, and another new subnet $\widehat{s^{(t)}} = \{o_{width}^{(t)}, {\bf o_{depth}^{ac}}\}$ by aligning all dimensions to anchor model except widths.  Then, we compute the FLOPs differences between two new subnets and the anchor model:
\begin{gather}
	\Phi_a (o_{width}^{(t)}) = {\rm FLOPs}(\overline{s^{(t)}}) - {\rm FLOPs}(s^{ac}), \\
	\Phi_b (o_{depth}^{(t)}) = {\rm FLOPs}(\widehat{s^{(t)}}) - {\rm FLOPs}(s^{ac}). 
\end{gather}
 If $\Phi_a\ge\Phi_b$,   subnet $s^{(t)}$ is considered to have wider widths and shallower depths for transformer stages, which adheres to path preference rule, and we will train it. Otherwise, we resample a new subnet and repeat the above steps. 

The quality of the anchor model $s^{ac}$ can impact the validity of these comparisons. We conjecture that the memory bank captures such preferences and thus select the subnet with minimal loss in the memory bank as the anchor model.

\subsection{Overall Supernet Training Process}
	\vspace{-1ex}
	Fig.~\ref{fig:alg} shows the supernet training process that uses both complexity-aware and performance-aware sampling techniques. At step $t$, we choose a FLOPs level $C^{(t)}_j$ from $\{C^{(t-1)}_{i-1}, C^{(t-1)}_{i}, C^{(t-1)}_{i+1}\}$ that is close to $C^{(t-1)}_i$  at step $t-1$. Then, we sample 1 smallest subnet and $M$=3 stochastic subnets for training. The smallest subnet is the nearest one from the HSS set with FLOPs closest to $C^{(t)}_j$. The $M$=3 subnets have FLOPs equal to $C^{(t)}_j$. They are either from the memory bank ${\rm B}_j$ with probability $q$ or based on the path preference rule with probability 1-$q$. We compute and accumulate the gradients for above 4 subnets. Then, we use the accumulated gradients to update supernet parameters.



%% file: eval.tex
\section{Evaluation}
	\vspace{-1ex}
\noindent\textbf{Setup}. We apply our proposed techniques to train our very large ViT supernet for 600 epochs. The complexity levels used in the training are set to \{100, 200, 300, 400, 500, 700, 900, 1200\} MFLOPs, which are suitable for mobile-regime ViTs. The other training setting and  hyper-parameters follows the existing best practices~\cite{nasvit,alphanet}. We list the detailed numbers in supplementary materials.

Then, we evaluate the effectiveness of our trained ViT supernet on four mobile phones with varying resource levels: weak (Pixel1), neutral (Pixel4, Xiaomi11), and strong (Pixel6). For each device, we set a range of latency constraints and use nn-Meter~\cite{nnmeter} to build a latency predictor for efficient search. We use the evolutionary search method in OFA~\cite{ofa} to search 5k subnets for each latency constraint.

\subsection{Main Results on ImageNet}
	\vspace{-1ex}
\begin{table}[t]
	\fontsize{8}{8} \selectfont
\caption{{\sysname} performance on ImageNet-1K~\cite{imagenet} with comparison to state-of-the-art efficient CNN and ViT models. We group models based on the hardware they are suited  according to FLOPs.   $^*$: latency is measured on each group's corresponding hardware.    }
\vspace{-3ex}
\label{tbl:overallresults}
\begin{tabular}
	{@{\hskip1pt}ccc@{\hskip2pt}c@{\hskip2pt}c@{\hskip1pt}}
\toprule
\multicolumn{5}{c}{\textbf{Tiny models:$<$100  MFLOPs for weak Pixel1 phone}}                                                                   \\ \hline
Model                        & MFLOPs        & Top-1 Acc                  &Latency$^*$       & Type                \\ \hline
ShuffleNet-V2 x0.5 \cite{ma2018shufflenet}            & 41            & 60.3              &        8.7 ms         & CNN                 \\
\textbf{ElasticViT-T0}                & \textbf{37}            & \textbf{61.1}                    &  \textbf{8.5 ms}        & ViT NAS          \\
MnasNet-x0.35 \cite{tan2019mnasnet}               & 63            & 64.1              &   13.5 ms             & CNN NAS             \\
MobileFormer~\cite{mobileformer}                 & 52            & 68.7               &       176.6 ms         & ViT              \\
\textbf{ElasticViT-T1}               & 62           &       67.2            &   \textbf{13.1 ms}        & ViT NAS          \\ \midrule
\multicolumn{5}{c}{\textbf{Tiny models: 100$\sim$150 MFLOPs for weak Pixel1 phone}  }                                                      \\ \hline
ShuffleNet-V2 1$\times$ \cite{ma2018shufflenet} &146&69.4& 24.3 ms &CNN\\
Cream~\cite{Cream}                        & 114           & 72.8                    &       24.3 ms    & CNN NAS             \\
FBNet-v2 \cite{wan2020fbnetv2}                      & 126           & 73.2                   &     22.5 ms       & CNN NAS             \\
MobileNet-V3 x0.75 \cite{mobilenetv3}            & 155           & 73.3              &     26.4 ms            & CNN NAS             \\
MobileFormer~\cite{mobileformer}                   & 96          & 72.8                    &        216.3 ms   & ViT              \\
\textbf{ElasticViT-T2}                & \textbf{119}          & \textbf{73.8}          &        \textbf{22.4 ms}       & ViT NAS          \\ 
\textbf{ElasticViT-T3}                & 160         & \textbf{75.2}                &     \textbf{29.2 ms}    & ViT NAS          \\ 
\midrule
\multicolumn{5}{c}{\textbf{Small models: 200$\sim$350 MFLOPs for neutral  Pixel4 phone}}                                                        \\ \hline
MobileNet-v3 x1.0 \cite{mobilenetv3}            & 219           & 75.2          &       24.2 ms              & CNN NAS             \\
FBNet-v2 \cite{wan2020fbnetv2}                    & 238           & 76.0              &           22.1 ms      & CNN NAS             \\
OFA \#25 \cite{ofa}                    & 230           & 76.4                  &      25.2 ms       & CNN NAS             \\
BigNAS-S \cite{bignas}                      & 242           & 76.5                   &      32.6 ms      & CNN NAS             \\
MobileFormer~\cite{mobileformer}                 & 214           & 76.7               &       415.1 ms         & ViT              \\
HR-NAS-A~\cite{ding2021hr}                   & 267           & 76.6                       &      -  & ViT NAS             \\
\textbf{ElasticViT-S1}       & \textbf{218} & \textbf{77.2}           &    \textbf{21.0 ms}       & ViT NAS \\
GreedyNAS-v2~\cite{greedynas}                 & 324           & 77.5               &          65.4 ms     & CNN NAS             \\

LeViT-128S~\cite{levit}                   & 305           & 76.6                 &       30.5 ms       & ViT             \\

HR-NAS-B~\cite{ding2021hr}                   & 325           & 77.3                  &    -         & ViT NAS             \\

\textbf{ElasticViT-S2} & \textbf{318} & \textbf{78.6}            &     \textbf{29.6 ms}    & \textbf{ViT NAS} \\ 
\midrule
\multicolumn{5}{c}{\textbf{Medium models: 350$\sim$500 MFLOPs for neutral Pixel4 phone}}                                                         \\ \hline

EfficientNet-B0 \cite{tan2019efficientnet}            & 390           & 77.1                &          55.1 ms     & CNN NAS             \\
BigNAS-M \cite{bignas}                  & 418           & 78.9                    &     64.3 ms      & CNN NAS             \\
MobileViT-XXS~\cite{mobilevit}& 364& 69.0 &84.1 ms&ViT\\
LeViT-128~\cite{levit}                    & 406           & 78.6                  &          40.2 ms   & ViT              \\

MobileViTv3-0.5~\cite{mobilevitv3}& 481& 72.3&96.7 ms&ViT\\
\textbf{ElasticViT-M} & \textbf{415} & \textbf{79.1}             &   \textbf{37.4 ms}     & ViT NAS \\ 
\midrule
\multicolumn{5}{c}{\textbf{Large models: $\geq$ 500 MFLOPs for  strong Pixel6 phone}}                                                         \\ \hline
MAGIC-AT~\cite{xu2022analyzing}                        & 598           & 76.8                &           37.9 ms    & CNN NAS             \\
BigNAS-L \cite{bignas}                  & 586           & 79.5                      &     45.7 ms    & CNN NAS             \\
MobileViTv2-0.5~\cite{mobilevitv2}                  & 500           & 70.2               &      56.5 ms          & ViT              \\
UniNet-B0~\cite{uninet}                      & 560           & 79.1                     &    53.9 ms      & ViT NAS              \\
\textbf{ElasticViT-L1} & \textbf{516} & \textbf{79.4}            &     \textbf{33.1 ms}    & ViT NAS \\

EfficientNet-B1 \cite{tan2019efficientnet}        & 700           & 79.1                    & 49.9 ms         & CNN NAS             \\
EdgeViT-XXS~\cite{edgevit}                     & 600           & 60.0                   &   69.6 ms         & ViT            \\
MobileViT-XS~\cite{mobilevit}&986&74.8&84.4 ms&ViT\\
Autoformer-Tiny~\cite{autoformer} & 1300 & 74.7&71.1 ms& ViT NAS\\
ViTAS-Twins-T~\cite{su2022vitas}& 1400&79.4& -&ViT NAS\\
\textbf{ElasticViT-L2} & \textbf{704} & \textbf{79.8}                 & \textbf{43.8 ms}   & ViT NAS \\ 
\textbf{ElasticViT-L3} & \textbf{806} & \textbf{80.0}                 &  \textbf{50.5 ms}  & ViT NAS \\ 
 \hline

\end{tabular}
\end{table}

\noindent\textbf{Comparison with efficient CNN and ViT models.} Table~\ref{tbl:overallresults} reports the comparison with state-of-the-art models including both strong CNNs and recent efficient ViTs.  Remarkably, for the first time, we utilize two-stage NAS to deploy lightweight and low-latency ViT models ranging from 37 to 800 MFLOPs, enabling us to bring accurate ViTs to a wide range of mobile devices. Without retraining or finetuning, our discovered subnets {\sysname} models significantly outperform all evaluated ViT and  CNN baselines.

First, \textit{our models significantly outperform prior ViTs that are designed for mobile devices.} 
 For tiny ViTs, our {\sysname}-T3 achieves 75.2\% accuracy under only 160 MFLOPs, which is 2.9\% better than MobileNetV3x0.75 in terms of similar FLOPs. For medium-sized ViTs, {\sysname}-M achieves 79.1\% accuracy under 415 MFLOPs, significantly outperforming existing mobile ViTs with 0.5\% and 4.8\% higher accuracy than LeViT and MobileViTv3~\cite{mobilevitv3}, respectively.  For large ViT models where ImageNet classification accuracy saturates, {\sysname}-L3 still has 0.6\% and 5.3\% accuracy improvement compared with Autoformer~\cite{autoformer} and ViTAS~\cite{su2022vitas} with 1.7$\times$ and 1.6$\times$ fewer FLOPs. Furthermore, our ViT models not only achieve high accuracy but also demonstrate fast real inference latency, making them practical for deployment on resource-constrained mobile phones. This sets them apart from other mobile ViT approaches that solely focus on reducing FLOPs. For instance, with only 214 MFLOP, MobileFormer~\cite{mobileformer} has a slow latency of 415.1 ms on Pixel4, which is 17.2$\times$ slower than MobileNetV3 and 19.7$\times$ slower than our {\sysname}-S1.
 

Second, \textit{our models also surpass lightweight CNNs with higher accuracy and lower latency.} For instance, {\sysname}-T0 achieves 61.1\% accuracy with only 37 MFLOPs, which outperforms ShuffleNetv2x0.5 with 0.8\% higher accuracy. {\sysname}-T3 achieves
the same accuracy of 75.2\% as MobileNetV3 with only 160 MFLOPs, while be 1.2$\times$ faster. This is the first time that ViT outperforms CNN with a faster speed on mobile devices within the 200 MFLOPs range.

\begin{figure*}[t]
	\centering
	\includegraphics[width=1\columnwidth]{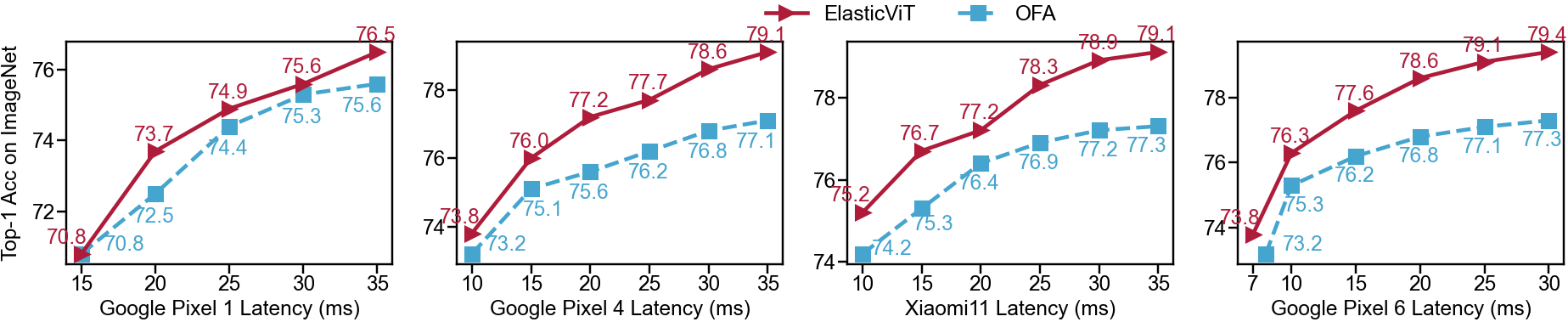}	
	\vspace{-4ex}
	\caption{Under the same latency constraint, the discovered ViTs by {\sysname} surpass state-of-the-art mobile CNNs on diverse mobile devices. From left to right: old and weak devices to latest and strong devices. }
	\label{fig:mobilelatency}
\end{figure*}



\noindent\textbf{Deploying efficient ViTs for diverse mobile phones}. Now we apply our ViT supernet to get different specialized subnets for diverse mobile devices. As a baseline for comparison, we choose OFA~\cite{ofa}, which represents the state-of-the-art hardware-aware NAS for delivering efficient mobile-regime CNNs. 
 Instead of relying solely on FLOPs to measure on-device efficiency, we use real inference latency for comparison. 
To ensure a fair comparison, we adopt the same approach as OFA and build a latency predictor for each of our test devices. We conduct an evolutionary search to find the optimal CNN subnets for each device. Note that we didn't compare with ViT NAS baselines, as existing works focus on large ViTs, whose supernets are out of range for searching low-latency models for mobile devices. 
 
Fig.~\ref{fig:mobilelatency} presents the latency-accuracy curve of our ViTs compared to OFA on diverse mobile devices. Our discovered ViT models consistently outperform OFA under various latency constraints, on both weak and strong mobile devices. Notably, our models are the first ViT models to achieve real-time inference latency on mobile devices without compromising accuracy. These results demonstrate the effectiveness of our approach and highlight the potential of transformers for efficient, high-performance models on mobile devices.

\subsection{Ablation Study}


We now conduct ablation studies to evaluate 1)   how each of our techniques can improve supernet training; 2) how our techniques mitigate the gradient conflict issues; and  3) the performance of  our searched model compared to retraining.

\begin{table}[t]
	\begin{center}
		\fontsize{8.5}{8.5} \selectfont
		\vspace{-3ex}
		\caption{Ablation study results on ImageNet. We show the top1 accuracy of best searched models for each case. Note that \textit{Multiple min} is applied on top of \textit{Adjacent sampling}; \textit{Perf-aware sampling} is applied on top of both \textit{Adjacent sampling} and \textit{Multiple min}.}
		\label{tbl:ablation_study}
		\begin{tabular}	
			{@{\hskip1pt}c@{\hskip3pt}c@{\hskip5pt}c@{\hskip5pt}c@{\hskip5pt}c@{\hskip5pt}c@{\hskip5pt}c@{\hskip5pt}c@{\hskip5pt}c@{\hskip1pt}}
			\midrule
			\multirow{2}{*}{Method} & \multicolumn{8}{c}{MFLOPs}\\ &100& 200 & 300 & 400& 500& 600 & 700 &800\\
			\midrule
			Baseline (Sandwich rule) &69.9 & 74.2& 76.5& 77.4&77.8&78.3&78.7&79.0 \\
			\midrule
			Adjacent step sampling &73.2 &75.6 &76.7 &77.5&78.2&78.3&78.7&79.0\\
			+Multiple min (HSS)&72.8 &76.7 &78.4 &79.0&79.3&79.4&79.6&79.7\\
			++Perf-aware sampling & \textbf{73.8}&\textbf{77.2} &\textbf{78.6} &\textbf{79.1}&\textbf{79.4}& \textbf{79.6}&\textbf{79.8} &\textbf{80.0}\\
			\midrule
		\end{tabular}   
	\end{center}
\end{table}

\vspace{2pt}
\noindent\textbf{Ablation study on each technique.} Table~\ref{tbl:ablation_study} shows the accuracy of the best-searched models under different supernet training techniques. We start with the baseline supernet that is trained using the original sandwich rule. Then we apply our techniques one by one to enhance the supernet training. We keep all the other training settings and search process consistent for a fair comparison. 

Table~\ref{tbl:ablation_study} illustrates the effectiveness of our proposed techniques in enhancing supernet training over a vast ViT search space. Both adjacent step sampling and multiple min strategy effectively controls the FLOPs differences among trained subnets,  resulting in substantial top-1 accuracy gains of up to 3.3\%. Furthermore, by further using performance-aware sampling to train good subnets, we are able to further improve the best-searched ViTs by up to 1\% accuracy.

\vspace{2pt}
\noindent\textbf{The effectiveness of mitigating gradient conflicts.} Our accuracy improvements primarily stem from the effective mitigation of gradient conflicts. To validate this, we conduct experiments on two supernets trained with different approaches: one using our proposed techniques and one using the sandwich rule as a baseline. We freeze the weights for both supernets and study the gradients of different subnets under the same batch of training data. We randomly sample 50 subnets under three levels of FLOPs: 50M, 200M, and 600M. We compute the cosine similarity of shared weights' gradient between each pair of subnets under the same FLOPs. A higher cosine similarity indicates less gradient conflict.

Table~\ref{tbl:ablation_study2} shows the average gradient similarity between subnets on both supernets. Compared to vanilla sandwich rule, we can significantly improve the gradient similarity for both random  and good subnets, suggesting that our method can efficiently mitigate the gradient conflicts.

\begin{table}[t]
	\begin{center}
		\small
		\fontsize{8.5}{8.5} \selectfont
		\caption{Gradient cosine similarity between subnets in supernet trained with different methods. The "good" subnets refer to the top10 subnets chosen from the randomly sampled 50 subnets.}
		\vspace{-3ex}
		\begin{tabular}	
			{@{\hskip2pt}c@{\hskip2pt}|c@{\hskip5pt}c@{\hskip5pt}|c@{\hskip5pt}c@{\hskip5pt}|c@{\hskip5pt}c@{\hskip1pt}}
			\hline
			\multirow{3}{*}{Method}&\multicolumn{6}{c}{MFLOPs}\\
			\cline{2-7}
			&\multicolumn{2}{c|}{50}&\multicolumn{2}{c|}{200}&\multicolumn{2}{c}{600}\\
			&Random&Good&Random&Good&Random&Good\\
			\hline
			Sandwich rule &0.37 & 0.47&0.32 &0.41 &0.31 & 0.46\\
			\textbf{Ours} &\textbf{0.50} & \textbf{0.56}& \textbf{0.51}&\textbf{0.67} &\textbf{0.51} &\textbf{0.67} \\
			\hline
		\end{tabular}   
		\label{tbl:ablation_study2}
	\end{center}
\end{table}

\noindent\textbf{Comparison with training from scratch}. High-quality supernet training can ensure that subnets achieve comparable accuracy as those trained from scratch. We retrain each subnet with a batch size of 512 on 8 Nvidia V100 GPUs, following the same training settings as LeViT~\cite{levit}. 
 Table~\ref{tbl:ablation_study1} compares the accuracy on ImageNet. These selected subnets can achieve even higher accuracy by directly inheriting weights from our supernet, with up to 2\% improvement. Interestingly, we notice that larger ViT models achieve comparable accuracy to retraining, while  tiny and small ViTs ($<$500 MFLOPs) can benefit more from supernet training.

\begin{table}[t]
	\begin{center}
		\small
		\fontsize{8.5}{8.5} \selectfont
			\caption{Best-searched ViT top-1 accuracy of inheriting supernet's weights vs.  retraining from scratch.}
			\vspace{-3ex}
		\begin{tabular}
		 {@{\hskip1pt}c@{\hskip7pt}c@{\hskip7pt}c@{\hskip7pt}c@{\hskip7pt}c@{\hskip7pt}c@{\hskip7pt}c@{\hskip7pt}c@{\hskip7pt}c@{\hskip7pt}c@{\hskip1pt}}
		 \hline		 \multirow{2}{*}{Method} & \multicolumn{8}{c}{MFLOPs}\\
		 &100& 200 & 300 & 400& 500& 600 & 700 & 800\\
		 \hline
		 Train from scratch &73.4 &75.3 & 76.6& 77.7& 79.0&79.5&\textbf{79.9}& 80.0\\
		  \textbf{\sysname}  &\textbf{73.8}& \textbf{77.2}&\textbf{78.6} &\textbf{79.1} &\textbf{79.4} &\textbf{79.6} &79.8&80.0\\
			\hline
		\end{tabular}   
		\label{tbl:ablation_study1}
	\end{center}
\end{table}

\vspace{-1ex}
\subsection{Transfer Learning}
\vspace{-1ex}
We transfer {\sysname} to  a list of commonly 
used transfer learning datasets: 1) CIFAR10 and CIFAR100~\cite{cifar10}; 2) fine-grained classification: Food-101~\cite{food101}, Oxford Flowers~\cite{flower} and Pets~\cite{pets}. We take the pretrained checkpoints on ImageNet and fine-tune on new datasets. We closely follow the hyper-parameter settings in GPipe~\cite{gpipe}.  The results are summarized in 
Table~\ref{tbl:finegrainedcls}. Compared to existing efficient CNNs and ViTs, our {\sysname} models achieves significantly better accuracy with fast inference speed on Pixel 4.

\begin{table}[t]
	\begin{center}
		\fontsize{8.2}{8.2} \selectfont
		\caption{Transfer learning results on downstream image classification datasets. We measure the latency on Pixel 4. }
		\vspace{-2.5ex}
		\begin{tabular}	
			{@{\hskip1pt}c@{\hskip4pt}c@{\hskip4pt}c@{\hskip4pt}c@{\hskip3pt}c@{\hskip3pt}c@{\hskip4pt}c@{\hskip4pt}c@{\hskip1pt}}
			\hline
		{Model}  &Latency& CIFAR10&CIFAR100&Food-101&Flowers&Pets\\
			\hline
			MobileNetV3&24.2 ms&97.1&83.3&86.8&94.3&87.7\\
			\textbf{{\sysname}-S1}&\textbf{21.0 ms}&\textbf{97.5}&\textbf{86.1}&\textbf{87.2}&\textbf{94.3}&\textbf{92.1}\\
			\hline
			LeViT-128S&30.5 ms&96.8&85.0&73.6&86.2&90.1\\
			\textbf{{\sysname}-S2}&\textbf{29.6 ms}&\textbf{97.5}&\textbf{86.9}&\textbf{88.3}&\textbf{95.2}&\textbf{92.9}\\
			\hline
			EfficientNet-B0&55.1 ms&97.9&86.9&89.1&92.4&92.2 \\
			LeViT-128&40.2 ms&97.8&86.6&80.8&86.2&92.2\\
			\textbf{{\sysname}-M}&\textbf{37.5 ms}&\textbf{97.9}&\textbf{87.0}&88.8&\textbf{95.6}&\textbf{93.3}\\
			\hline
		\end{tabular}   
		\label{tbl:finegrainedcls}
	\end{center}
\end{table}

%% file: conclusion.tex
	\vspace{-1ex}
\section{Conclusion}
	\vspace{-1ex}
In this paper, we propose {\sysname}, a two-stage NAS approach that trains a high-quality supernet for deploying accurate and low-latency vision transformers on diverse mobile devices. Our approach introduces two key techniques to address the gradient conflicts issue by constraining FLOPs differences among sampled subnets and sampling potentially good subnets, greatly improving supernet training quality.  Our discovered ViT models outperfom prior-art efficient CNNs and ViTs on the ImageNet dataset, establishing new SOTA accuracy under various of latency constraints. 